\documentclass{article}

\usepackage[final]{neurips_2023}

\usepackage{bm}      

\usepackage{amsmath, amssymb, amsfonts}
\usepackage[ruled,linesnumbered,algo2e]{algorithm2e}
\usepackage{mathtools}
\usepackage{graphicx}
\usepackage{subcaption}
\usepackage{booktabs}
\usepackage{algorithm}
\usepackage{algorithmic}
\usepackage{xcolor}
\usepackage{tikz}
\usetikzlibrary{positioning, arrows.meta, shapes.geometric}

\title{Enhancing Multi-Image Question Answering via Submodular Subset Selection}

\author{
  Aaryan Sharma$^1$, 
  Shivansh Gupta$^1$, 
  Samar Agarwal$^1$, 
  Vishak Prasad C.$^1$, 
  Ganesh Ramakrishnan$^1$ \\
  $^1$Indian Institute of Technology Bombay \\
}

\begin{document}

\maketitle

\begin{abstract}
Large multimodal models (LMMs) have achieved high performance in vision-language tasks involving single image but they can't handle when presented with a collection of multiple images (Multiple Image Question Answering scenario). These tasks,  which involve reasoning over large number of images, present issues in scalability (with increasing number of images) and retrieval performance. In this work, we propose an enhancement for retriever framework introduced in MIRAGE model using submodular subset selection techniques. Our method leverages query-aware submodular functions, such as GraphCut, Facility Location, and Log Determinant, to pre-select a subset of semantically relevant images before main retrieval component. We demonstrate that using anchor-based queries and augmenting the data improves submodular+retriever effectiveness, particularly in large haystack sizes. 

\end{abstract}

\section{Problem Statement}
Large Multimodal Models (LMMs) have significantly advanced vision-language understanding, particularly in single-image question answering. However, their capabilities remain constrained when applied to real-world tasks that involve reasoning over extensive collections of visual data, such as searching through photo albums or analyzing textual images. These large-scale multi-image question answering (MIQA) scenarios present unique computational and retrieval challenges that most LMMs are currently unable to handle efficiently.

The MIRAGE framework \cite{wu2025visual} addresses part of this gap by introducing a retrieval-augmented generation (RAG) approach primarily for vision tasks. MIRAGE outperforms proprietary models for large haystacks ($>$50 images) and can handle upto 10k images while proprietary models can handle only upto 100 images.  But, there is a critical limitation in the retriever component of the system. As the size of the image haystack increases, MIRAGE's retriever \cite{wu2025visual} suffers from reduced effectiveness in identifying relevant images. This inefficiency becomes particularly problematic in the haystacks of thousands of images, leading to low retriever accuracy and scalability issues. This performance degradation propagates when passed through the pipeline.

Although RAG-based methods have proven successful in NLP and have been adapted to vision language models using components such as CLIP-based encoders and BERT-RCNN hybrids, current retrievers, including MIRAGE \cite{wu2025visual}, lack the robustness required for fine-grained, large-scale visual selection. These systems still struggle to balance retrieval accuracy with runtime efficiency, especially when visual distractors are numerous.

Addressing this retriever bottleneck is critical for enabling scalable, high-performance MIQA. Improving retriever efficiency would not only enhance model accuracy but also reduce computational costs, making advanced vision-language models more practical for large-scale applications.

\section{Related Works}
The intersection of vision and language has significantly progressed in recent years, driven primarily by advanced multimodal models (LMMs). Despite these developments, scaling such models to handle multi-image inputs—especially in real-world settings—introduces key challenges, including effective image retrieval, memory constraints, and the need for coherent reasoning across diverse visual contexts. Recent research has explored both retrieval-augmented generation frameworks and intelligent data selection strategies to address these limitations. These efforts aim to enhance model efficiency and scalability while maintaining high performance in complex vision-language tasks. 

One influential framework in guided subset selection is PRISM \cite{kothawade2022prismrichclassparameterized}, which introduces a rich class of \textit{Parameterized Submodular Information Measures} for tasks such as guided summarization and targeted learning. Using instantiations of Mutual Information (MI), Conditional Gain (CG), and Conditional Mutual Information (CMI) on data, query and/or target set, PRISM \cite{kothawade2022prismrichclassparameterized} offers a general framework by enabling trade-offs between diversity, query relevance, query coverage, and privacy irrelevance. Submodular functions such as GraphCut, Facility Location and LogDeterminant helps in subset selection by enabling control over the selected subset in terms of query and privacy set.  

One recent line of work explores submodular-based methods for reducing training cost and improving scalability.
ORIENT \citep{NEURIPS2022_ce9440b9} proposes a Submodular Mutual Information (SMI)-based framework for supervised domain adaptation under distribution shift. It dynamically selects a representative and diverse subset of training data from a large source domain by leveraging a small, labeled reference set from the target domain. This targeted subset selection significantly reduces training time and resource usage while maintaining or improving task accuracy. Notably, ORIENT \citep{NEURIPS2022_ce9440b9} integrates well with existing training frameworks, making it a flexible and effective solution for data-efficient learning. This approach supports our goal by demonstrating the effectiveness of submodular selection in identifying compact, relevant subsets from large datasets. It validates our strategy to improve retriever efficiency by filtering visual haystacks prior to retrieval.

The contextual similarity framework proposed by Liao et al. \cite{liao2023supervisedmetriclearningrank} introduces a novel \textit{contextual loss} that learns an embedding space optimizing the semantic relationship rather than using a traditional pairwise nearest neighbour ranking approach that solely rely on cosine similarity. This approach is more robust to label noise and could help in retrieval tasks based on similarity.

The idea of learning submodular functions through neural networks emerges from an approach by Manupriya et al., who introduced the SEA-NN framework \cite{manupriya2022improvingattributionmethodslearning} for improving attribution methods using deep submodular functions. In our future work, we would adapt this strategy to select contextually important and diverse image subsets by using gradients from a QA loss to train a deep submodular function. This approach allows the retriever to benefit from data-aware, task-specific scoring, instead of relying solely on general-purpose submodular functions mentioned in PRISM \cite{kothawade2022prismrichclassparameterized}. 

\section{Approach of Solving}

Our solution tackles the MIQA challenge faced by Mirage model \cite{wu2025visual} by focusing on solving the retriever inaccuracy in selecting the needle image from larger haystacks. We propose a dual-strategy framework combining submodular subset selection \cite{kothawade2022prismrichclassparameterized} with alreay present retriever. 

\subsection{Subset Selection using Submodular Functions}

Instead of relying solely on Anchor or Target phrases from the input query, we propose using reference images to construct a query set. These images provide richer context for identifying semantically aligned images in the haystack.Inspired by the Prism framework \cite{kothawade2022prismrichclassparameterized}, we use a submodular approach to select a subset of the haystack by balancing query relevance ,diversity ,redundancy and privacy. 
\begin{itemize}
    \item \textbf{Submodular Functions}: We use the following functions \cite{kothawade2022prismrichclassparameterized}:
    \begin{itemize}
        \item GraphCut (GC): Prefers relevance and have a tradeoff between diversity and representativeness 
        \item Facility Location (FL): Favours diversity and query coverage
        \item Log Determinant (LogDet): Tradeoff between query-relevance and diversity 
    \end{itemize}
    By using these submodular functions, we aim to increase the retriever accuracy using Query-Preserving and Privacy-Preserving methods.  

    \item \textbf{Mixture of Submodular Functions}: We use a convex combination of submodular functions (Example- 70\% GC, 20\% FL, 10\%LD).

    \item \textbf{Deep Submodular Function}: Inspired from SEA-NN \cite{manupriya2022improvingattributionmethodslearning}, we aim to train a neural network to provide relevance score guided by gradients from QA loss. This helps to adapt to the dataset and query context instead of generalized submodular functions.
    
\end{itemize}
    

\subsubsection*{Data Augmentation for Query Set Expansion}
To increase the robustness of the query set, we used augmentation techniques such as:
\begin{itemize}
    \item Random Crop
    \item Random Flip
    \item Color Distortion (Jitter + Greyscale)
    \item Gaussian Blur
\end{itemize}






\subsection{Integration with Mirage Framework}
We integrate the subset selection into the Mirage framework \cite{wu2025visual} to have a more enhanced retriever for efficient retrieval by pre-filtering (submodular functions). 


\section{Experiments and Results}

\begin{figure}[htbp]
    \centering
    \includegraphics[width=0.7\linewidth]{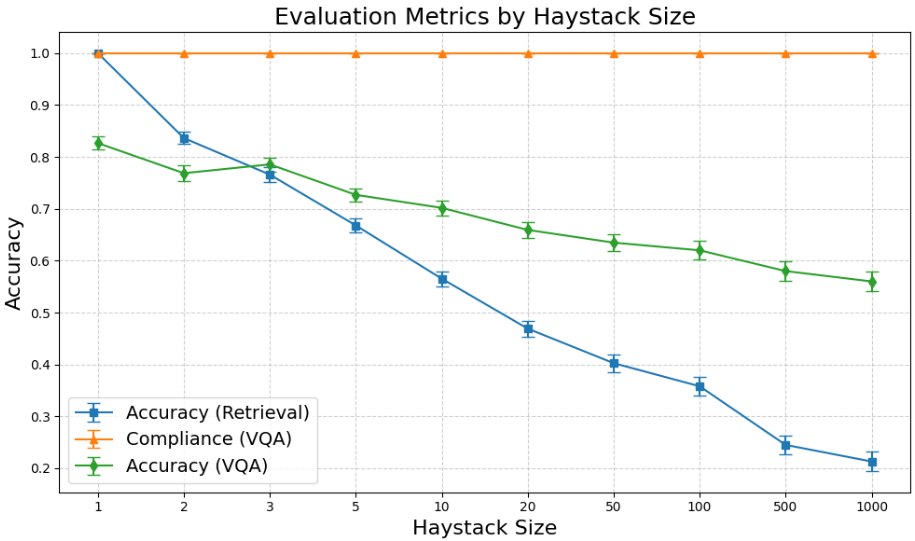}
    \caption{Evaluation of MIRAGE and Retriever}
    \label{fig:eval_metrics}
\end{figure}

This section presents our experimental setup and findings aimed at improving the performance of the MIRAGE model, introduced in the Visual Haystacks paper~\cite{wu2025visual}. Our primary objective is to enhance the effectiveness of the Retriever module, particularly in scenarios involving varying haystack sizes. Here, \textit{haystack size} refers to the number of images provided to the model in a single retrieval instance. As anticipated, we observed a decline in retrieval accuracy with increasing haystack sizes as evident in Figure~\ref{fig:eval_metrics}.

\begin{figure}[htbp]
    \centering
    \includegraphics[width=0.5\linewidth]{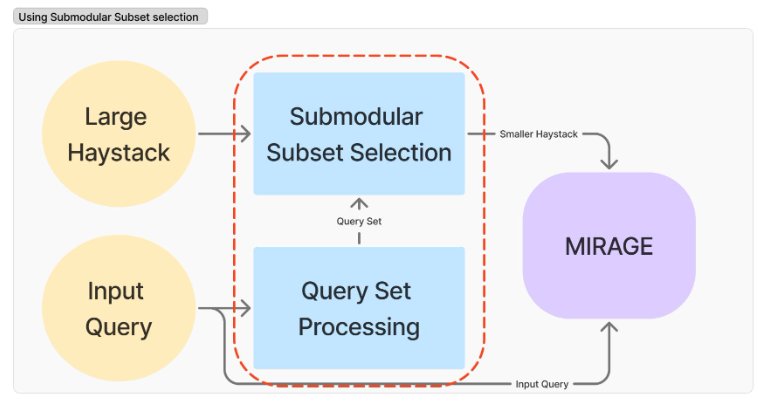}
    \caption{Pipeline for subset selection}
    \label{fig:block_diagram}
\end{figure}

To address this limitation, we propose a pre-selection mechanism that filters the haystack down to a smaller, more relevant subset of images. This subset is then passed to the Retriever, increasing the probability of successful retrieval by narrowing the search space. Block diagram for the pipeline is presented in Figure~\ref{fig:block_diagram}. We first perform query set processing using input query,we retrieve a query image from a set containing reference image for all classes to form the query set. Our subset selection strategy is inspired by submodular optimization techniques, particularly those described in the PRISM framework~\cite{kothawade2022prismrichclassparameterized}.

The Visual Haystacks benchmark uses query templates of the form: \emph{``For the image with an \{Anchor\}, is there a \{Target\}?”} For example: \emph{``For the image with a Truck, is there a Dog?”}.  We explored both the Anchor and Target components of the query as potential inputs for the subset selection module. All experiments were conducted on the COCO dataset~\cite{cocodataset}. Notably, this VHS benchmark ~\cite{wu2025visual} yields a binary output (\emph{Yes/No}) for each query. In contrast, a more general setting MIQA task could involve open-ended or more complex queries such as \emph{"Which image best captures the setting of a sunset?"}

For subset selection, we employed a combination of submodular functions such as Facility Location Mutual Information (FLVMI), Graph Cut Mutual Information (GCMI), and Log Determinant (LogDet). These functions measure the informativeness of candidate images with respect to the query, using cosine similarity as the similarity metric. A NaiveGreedy optimizer was used to select the top-ranked images. Further details on these methods can be found in~\cite{kothawade2022prismrichclassparameterized}.

\subsection{Subset Selection}

We integrated submodular subset selection into the MIRAGE retrieval pipeline to identify relevant image subsets prior to retrieval. The process computes pairwise cosine similarity between the features of a query image (either Anchor or Target) and the dataset extracted using VGG network, selecting images based on the chosen submodular optimization criteria.

\begin{figure}[htbp]
  \centering
  \includegraphics[scale = 0.5]
  {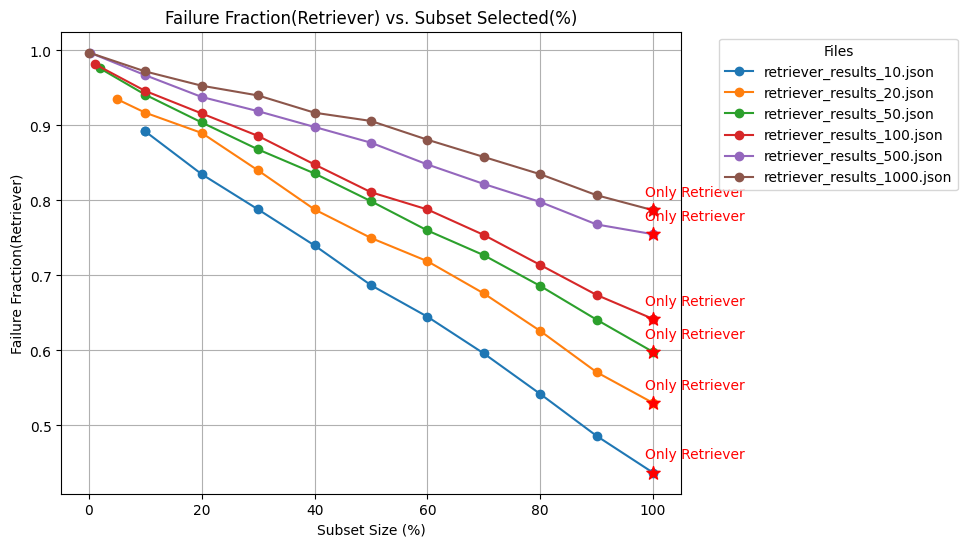}
  \caption{Retriever performance using Target-based subset selection. The x-axis represents the subset size as a percentage of the total haystack. Different legends are for different size of haystacks.}
  \label{fig:wide_figure}
\end{figure}

In our initial experiments, the Target image was used as the query input, and the GCMI function was used for subset selection. The filtered subset was then passed to the Retriever to identify the ground-truth (needle) image. As shown in Figure~\ref{fig:wide_figure}, this approach did not yield improved performance compared to directly using the entire haystack, suggesting that Target-based selection may not effectively capture contextual relevance.

To better understand the lack of performance improvement, we evaluated the effectiveness of the submodular selection process itself and a significant portion of the selected subsets did not include the ground-truth image, which inherently limits the Retriever’s ability to succeed.

\begin{figure}[htbp]
  \centering
  \includegraphics[width=0.9\textwidth]{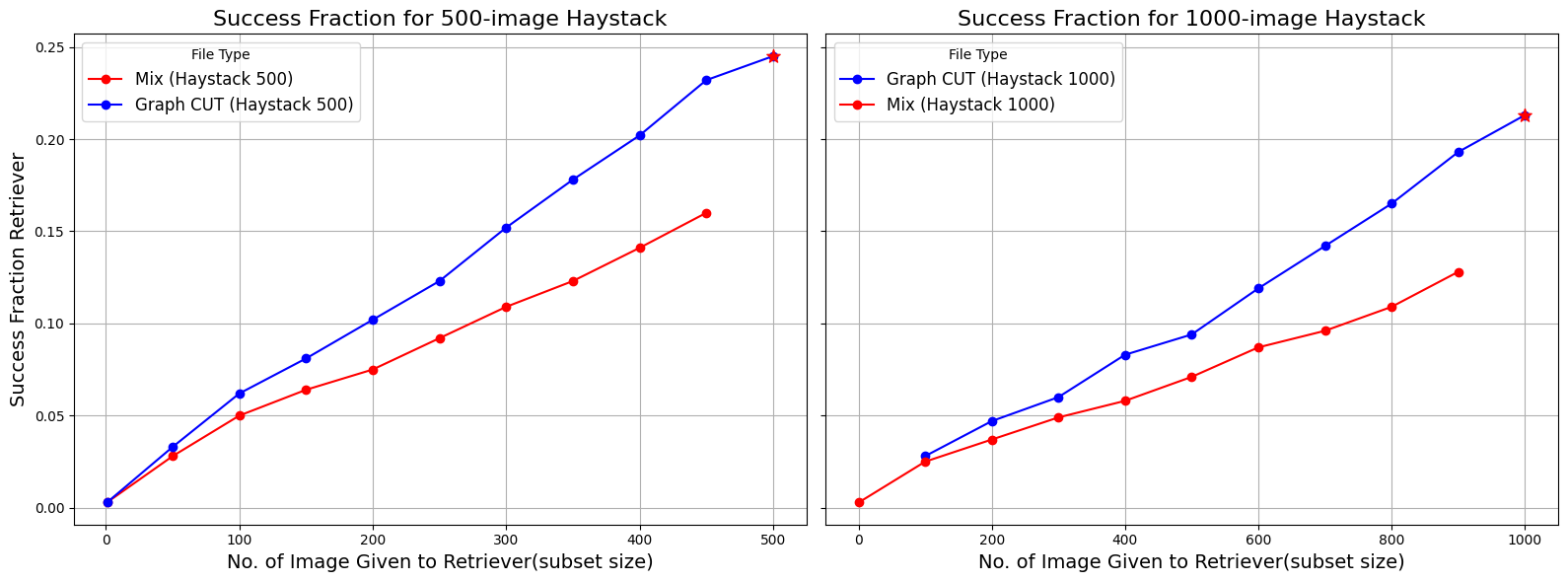}
  \caption{Performance comparison of single-function (GCMI) vs. mixed-function (GCMI, FLVMI, LogDet) subset selection for haystack sizes of 500 and 1000.}
  \label{fig:MIx vs GCMI}
\end{figure}

To improve the subset quality, we experimented with a hybrid approach combining GCMI, FLVMI, and LogDet in a 70:20:10 ratio. However, as shown in Figure~\ref{fig:MIx vs GCMI}, the mixed-function strategy did not outperform the single-function GCMI method, and in fact, yielded slightly poorer results.

\begin{figure}[htbp]
  \centering
  \includegraphics[width=0.8\textwidth]{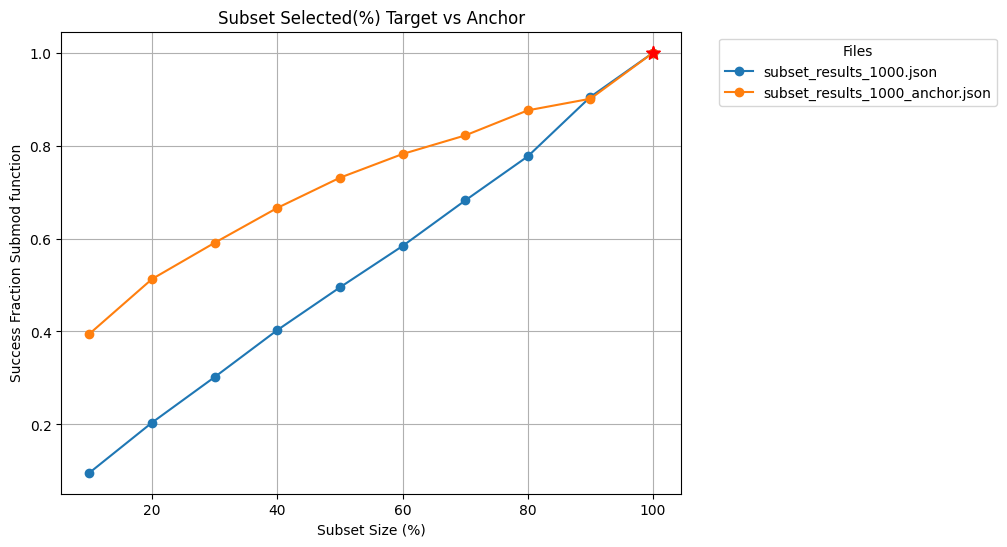}
  \caption{Comparison of subset selection using Anchor vs. Target images as queries.}
  \label{fig:anchor_target}
\end{figure}

Following these observations, we conducted a deeper analysis of the COCO dataset and random retrieval outputs, which revealed that Anchor images provide stronger contextual cues than Target images. Motivated by this insight, we modified our approach to use Anchor images as the query input for subset selection. This revised experiment was conducted with a haystack size of 1000 images. A comparative evaluation between Anchor- and Target-based selection strategies is presented in Figure~\ref{fig:anchor_target}.

These results demonstrate that using Anchor images for subset selection significantly increases the probability of the ground-truth image being included in the selected subset. Consequently, this improves the overall retrieval performance of the MIRAGE model by providing the Retriever with a more semantically aligned search space. Based on these findings, we now focus primarily on Anchor-based subset selection in subsequent experiments.

\begin{figure}[htbp]
    \centering
    \includegraphics[width=0.8\textwidth]{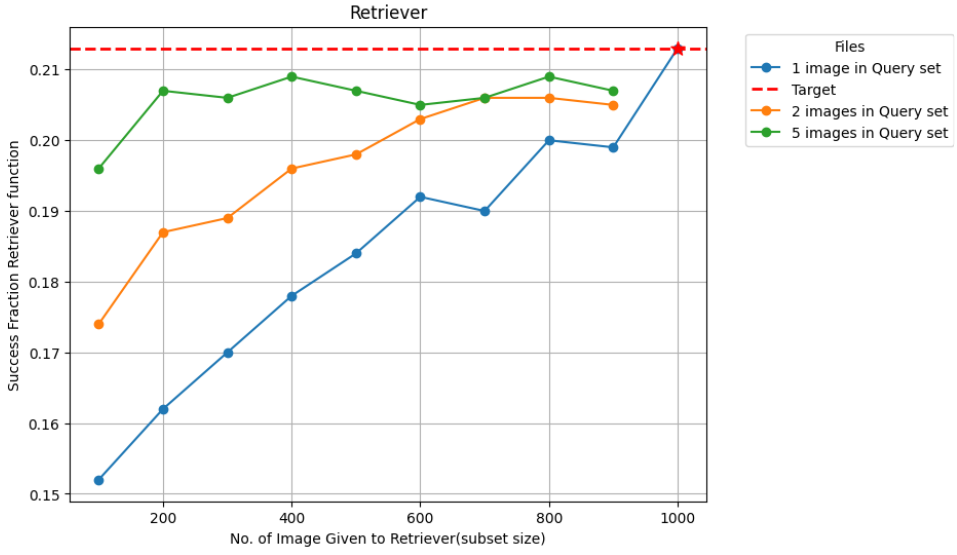}
    \caption{Increasing the query set size by adding more reference images}
    \label{fig:query_size}
\end{figure}

Next, instead of having only one reference image we increase the query set to 2 and 5 images by forming a fixed set from the COCO dataset for each class. We notice a significant increase in retriever success fraction (Figure~\ref{fig:query_size}) because of the fact that COCO dataset images contains various objects in a single image and computing similarity based on lesser number of images can capture high similarity of other objects as well. 

\begin{figure}[htbp]
    \centering
    \includegraphics[width=0.8\textwidth]{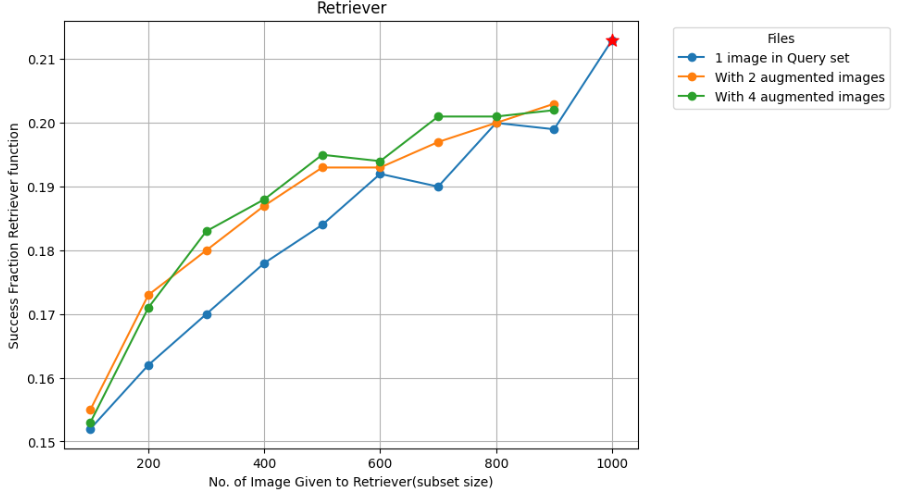}
    \caption{Success fraction for data augmented query set}
    \label{fig:data_augmentation}
\end{figure}

To increase the robustness of the query set, we used augmentation techniques mentioned in SimCLR \cite{chen2020simpleframeworkcontrastivelearning}. We apply four types of augmentation (random crop, random flip, color distortion and gaussian blur) each with a distinct randomness probability mentioned in SIMCLR Appendix A \cite{chen2020simpleframeworkcontrastivelearning}. We observe an increase in retriever accuracy with augmented images as evident in Figure~\ref{fig:data_augmentation} (but there is very slight difference between adding 2 and 4 augmented images which may suggest saturation).

\subsection{Analyzing the Retriever}

To further understand the behavior of the Retriever module in the MIRAGE pipeline, we analyzed its predictions across a diverse set of queries. Our observations can be categorized into three primary types of failure cases, each revealing different challenges in the retrieval process:

\begin{enumerate}
    \item \textbf{Anchor Confusion:} The Retriever often selects images that contain objects visually or semantically similar to the Anchor object, even when they are incorrect. For example, in one case, an image with a snowboard was retrieved instead of the correct image with a skateboard.
    \item \textbf{Target Salience Bias:} When the Target object is more prominently visible in a distractor image than in the correct image (where the Anchor is less visible), the Retriever tends to favor the distractor.
    \item \textbf{Anchor Ambiguity:} When multiple images in the haystack contain the Anchor object but only one matches the full query, the Retriever sometimes fails to resolve this ambiguity and selects the wrong image.
\end{enumerate}
\begin{figure}[H]
    \centering
    \begin{minipage}{0.45\textwidth}
        \includegraphics[width=\textwidth]{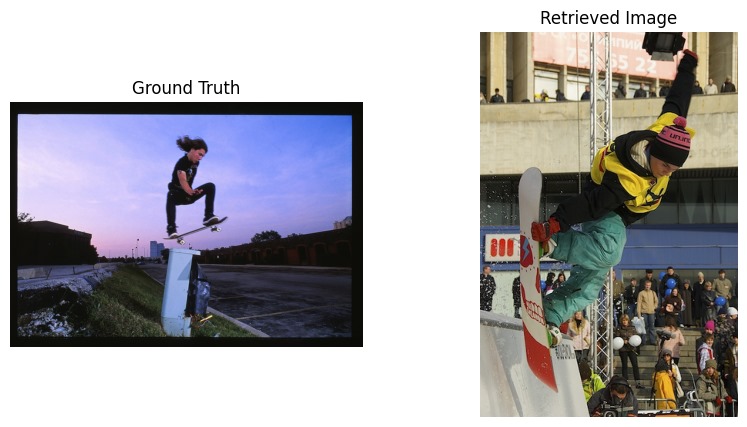}
    \end{minipage}
    \hfill
    \begin{minipage}{0.5\textwidth}
        \vspace{1.2cm}
        \raggedright
        \scriptsize
        \textbf{1.} Image with similar Anchor selected (e.g., snowboard instead of skateboard) \\ 
        \textbf{Query: For the image with a skateboard, is there a bed?}
    \end{minipage}
    
    \vspace{0.5cm}
    
    \begin{minipage}{0.45\textwidth}
        \includegraphics[width=\textwidth]{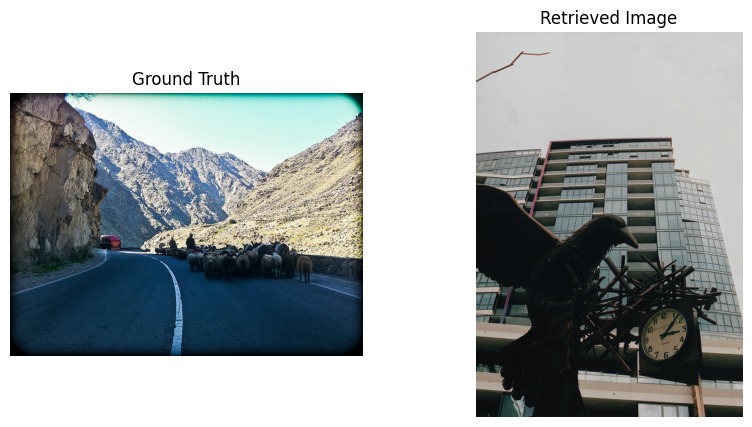}
    \end{minipage}
    \hfill
    \begin{minipage}{0.5\textwidth}
        \vspace{1.2cm}
        \raggedright
        \scriptsize
        \textbf{2.} Target more visible in distractor; Anchor is less visible in ground truth \\
        \textbf{Query: For the image with a cow, is there a clock?}
    \end{minipage}
    
    \vspace{0.5cm}
    
    \begin{minipage}{0.45\textwidth}
        \includegraphics[width=\textwidth]{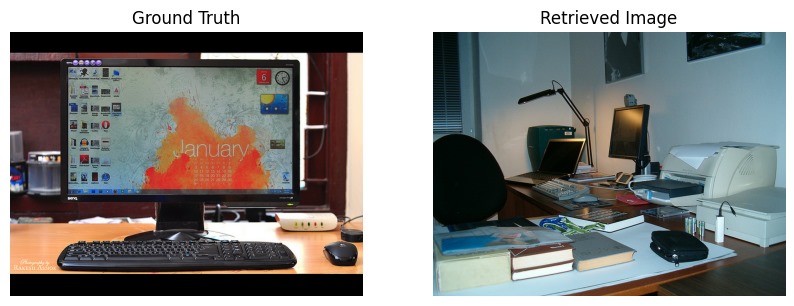}
    \end{minipage}
    \hfill
    \begin{minipage}{0.5\textwidth}
        \vspace{1.2cm}
        \raggedright
        \scriptsize
        \textbf{3.} Multiple images contain Anchor; Retriever confuses the correct needle \\
        \textbf{Query: For the image with a tv, is there a car?}
    \end{minipage}
    
    \caption{Failure cases observed in Retriever outputs across different query types. Each row highlights a specific issue affecting retrieval accuracy.}
    \label{fig:retriever_analysis}
\end{figure}

\section{Future Works}
As an extension to the MIQA task, future work will explore Document Haystacks \cite{chen2024documenthaystacksvisionlanguagereasoning}, where reasoning is performed over a pile of visual documents. These scenarios involve not only image understanding but also structured document processing,requiring integration of OCR, visual layout analysis and caption generation.

\bibliographystyle{unsrt}
\bibliography{references}

\end{document}